\documentclass[runningheads]{llncs}
\usepackage{graphicx}
\usepackage{multirow}

\usepackage{amsfonts}
\usepackage{subfig}
\usepackage{amsmath}
\usepackage{textcomp}
\usepackage[ruled]{algorithm2e} % For algorithms

 \newlength\mylen
\newcommand\myinput[1]{%
  \settowidth\mylen{\KwIn{}}%
  \setlength\hangindent{\mylen}%
  \hspace*{\mylen}#1\\}

\SetKwInOut{Parameter}{parameter} 
 
\usepackage{makecell}  
\begin{document}

\title{Learning Discriminative Features using Multi-label Dual Space} 
\titlerunning{Learning Discriminative Features using Multi-label Dual Space}
%authors list
\author{Ali Braytee \and
Wei Liu} %\and
%%Third Author\inst{3}\orcidID{2222--3333-4444-5555}}
%
\authorrunning{A. Braytee and W. Liu}
\institute{School of Computer Science \\
University of Technology Sydney, Sydney, Australia\\
\email{\{ali.braytee,wei.liu\}@uts.edu.au}}
% \and
%%Springer Heidelberg, Tiergartenstr. 17, 69121 Heidelberg, Germany
%\email{\{ali.braytee,wei.liu\}@uts.edu.au}%\\
%%\url{http://www.springer.com/gp/computer-science/lncs} \and
%%ABC Institute, Rupert-Karls-University Heidelberg, Heidelberg, Germany\\
%%\email{\{abc,lncs\}@uni-heidelberg.de}}
%
\maketitle              % typeset the header of the contribution
\begin{abstract}
Multi-label learning handles instances associated with multiple class labels. The original label space is a logical matrix with entries from the Boolean domain $\in \left \{ 0,1 \right \}$. Logical labels are not able to show the relative importance of each semantic label to the instances. The vast majority of existing methods map the input features to the label space using linear projections with taking into consideration the label dependencies using logical label matrix. However, the discriminative features are learned using one-way projection from the feature representation of an instance into a logical label space. Given that there is no manifold in the learning space of logical labels, which limits the potential of learned models. In this work, inspired from a real-world example in image annotation to reconstruct an image from the label importance and feature weights. We propose a novel method in multi-label learning to learn the projection matrix from the feature space to semantic label space and projects it back to the original feature space using encoder-decoder deep learning architecture. The key intuition which guides our method is that the discriminative features are identified due to map the features back and forth using two linear projections. To the best of our knowledge, this is one of the first attempts to study the ability to reconstruct the original features from the label manifold in multi-label learning. We show that the learned projection matrix identifies a subset of discriminative features across multiple semantic labels. Extensive experiments on real-world datasets show the superiority of the proposed method. 
 
\end{abstract}
\section{Introduction}
\label{intro}
%jian2016multi
Multi-label learning deals with the problem where each sample is represented by a feature vector and is associated with multiple concepts or semantic labels. For example, in image annotation, an image may be annotated with different scenes; or in text categorization, a document may be tagged to multiple topics. Formally, given a data matrix $X \in \mathbb{R}^{d \times n}$ is composed of $n$ samples of $d$-dimensional feature space. The feature vector $x_{i} \in X$ is associated with label set $Y_{i}=\left \{ y_{i1},y_{i2},_{\cdots},y_{ik}  \right \}$ where $k$ is the number of labels, and $y_{(i,j)} \in \left \{ 0,1 \right \}$ is a logical value where the associated label is relevant to the instance $x_{i}$. Over the past decade, many strategies have been proposed in the literature to learn from multi-labeled data. Initially, the problem was tackled by learning binary classification models on each label independently~\cite{zhang2013review}. However, this strategy ignores the existence of label correlation. Interestingly, several methods~\cite{braytee2019correlated,wang2019discriminative,wang2020dual} show the importance of considering the label correlation during multi-label learning to improve the classification performance. However, these methods use logical labels where no manifold exist and apply traditional similarity metrics such as Euclidean distance which is mainly built for continuous data.

\begin{figure}[t]
\centering
\subfloat[]{
\includegraphics[height=0.1\textheight,width=0.3\textwidth]{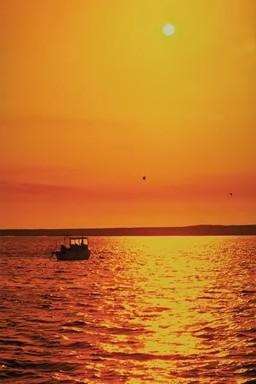}
\label{fig:boat1}}
%\hfill
\subfloat[]{
\includegraphics[height=0.1\textheight,width=0.3\textwidth]{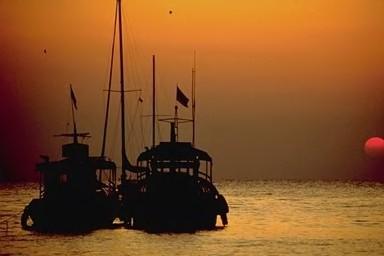}
\label{fig:boat2}}
\caption{Two images annotated with labels: sea, sunset, and boat}
\label{fig:natural}
\end{figure}

In this study, contrary to the majority of the methods, in addition to learning the mapping function from a feature space to multi-label space, we explore the projection function from label space to feature space to reconstruct the original feature representations. For example, in image annotation, our novel method is able to reconstruct the scene image using the projection function and the semantic labels. Initially, it is necessary to explore the natural structure of the label space in multi-labeled data. Existing datasets naturally contain logical label vectors which indicate whether the instance is relevant or not relevant to a specific label. For example, as shown in Fig.~\ref{fig:natural}, both images tagged the label boat with the same weight equal to $1$ (present). However, to accurately describe the labels in both images, we need to identify the importance of the labels in each image. It is clearly seen that the label boat in image (\ref{fig:boat2}) is more important than that in image (\ref{fig:boat1}). Furthermore, the label with the zero value in the logical label vector refers to different meanings, which may either be irrelevant, unrepresented or missing. Using the same example in Fig.~\ref{fig:natural}, the boat and the sun labels in images (\ref{fig:boat1} and \ref{fig:boat2}) are not tagged due to their small contribution (unrepresented). Our method learns a numerical multi-label matrix in semantic embedding space during the optimization method based on label dependencies. Therefore, replacing the importance of labels using numerical values instead of the logical labels can improve the multi-label learning process.   

Importantly, learning the numerical labels is essential to our novel approach that is developed based on the encoder-decoder deep learning paradigm~\cite{ranzato2007unified}. Specifically, the input training data in the feature space is projected into the learned semantic label space (label manifold) as an encoder step. In this step, simultaneously through an optimization problem, it learns the projection function and the semantic labels in Euclidean space. Significantly, we also consider the reconstruction task of the original feature representations using the projection matrix as input to a decoder. This step imposes a constraint to ensure that the projection matrix preserves all the information in the original feature matrix. The decoder allows the original features to be recovered using the projection matrix and the learned semantic labels. In the case of image annotation, this process is similar to combining puzzle pieces to create the picture. However, in the case of logical labels where the label either exists or not, it is incapable of reconstructing the original visual feature representations. We show that the impact of the decoder in identifying the relevant features can improve multi-label classification performance. This is because the feature coefficients are estimated based on the actual numerical labels and more importantly the weights of the relevant features results in a reduction of the reconstruction error. The proposed method is visualized in Fig.~\ref{fig:correlation}. We test the proposed approach on a variety of public multi-label datasets, and clearly verify that they favourably outperform the state-of-the-art methods in feature selection and data reconstruction.

We formulate the proposed approach as a constrained optimization problem to project feature representations into semantic labels with a reconstruction constraint. More precisely, the method is designed by an effective formulation of the encoder and decoder model using a linear projection to and from the learned semantic labels, respectively. This design alleviates the computational complexity of the proposed approach making it suitable for large scale datasets. To the best of our knowledge, this is the first attempt to learn the semantic label representation from the training data that can be used for data reconstruction in multi-label learning. In summary our contribution are: (1) a semantic encoder-decoder model to learn the projection matrix from original features to semantic labels that can be used for data reconstruction; (2) we extend the logical label to a numerical label which describes the relative importance of the label in a specific instance; (3) we propose a novel \textit{Learning Discriminative Features using Multi-label Dual Space (LDFM)} which is able to identify discriminative features across multiple class labels. 

\section{Related Work}
%nguyen2019multi
\textbf{Label correlation} \hspace{0.2cm} Over the past decade until recently, it has been proven that label correlation has improved the performance of multi-label learning methods. The correlation is considered either between pairs of class labels or between all the class labels which is known as second and high order approaches, respectively~\cite{zhang2013review,zhu2017multi,che2020novel}. However, in these models, the common learning strategy is to deal with logical labels which represents whether the label is relevant or irrelevant to an instance. The label matrix in the available multi-labeled datasets contains logical values which lack semantic information. Hence, few works reveal that transforming the labels from logical into numerical values improves the learning process.

\textbf{Semantic labels} \hspace{0.2cm} The numerical value in the label space carries semantic information, i.e., the value may refer to the importance or the weight of an object in the image. The numerical label matrix in Euclidean space is not explicitly available in the multi-labeled data. A few works have studied the multi-label manifold by transforming the logical label space to the Euclidean label space. For example, \cite{hou2016multi} explore the label manifold in multi-label learning and reconstruct the numerical label matrix using the instance smoothness assumption. Another work~\cite{cai2018multi} incorporates feature manifold learning in the multi-label feature selection method, and \cite{huang2018manifold} select the meaningful features using the constraint Laplacian score in manifold learning. However, our proposed method differs from these by learning an encoder-decoder network to reconstruct the input data using the learned projection matrix along with predicting the semantic labels. %of the instances.

\textbf{Autoencoder} \hspace{0.2cm} Several variants use an autoencoder for multi-label learning. \cite{cheng2019multi} learn the unknown labels using the entropy measure from existing labels, then the completed label matrix is used as an input layer feature set in autoencoder architecture. However, our method reconstructs the original input data in the decoder using the learned semantic labels. Further, \cite{law2019multi} propose a stacked autoencoder for feature encoding and an extreme learning machine to improve the prediction capability. However, the authors did not take label correlation into consideration and the original logical labels are used in the learning process. In this paper, we select the discrete features that are important to detect the objects' weights during the encoding phase and simultaneously they are significant to reconstruct the original data in the decoding phase.   
%\cite{lian2019captured} use a supervised autoencoder to learn the data distribution of the training samples and they incorporate a partial label dependence using the original logical labels. Finally

\section{The Proposed Method}
%\subsection{Preliminaries}
In multi-label learning, as mentioned above, the training set of multi-labeled data can be represented by $\left \{ x_{i} \in X | i=1,\cdots,n \right \}$, the instance $x_{i} \in \mathbb{R}^{d}$ is a $d$-dimensional feature vector associated with the logical label $Y_{i}=\left \{ y_{i1},y_{i2},_{\cdots},y_{ik} \right \}$, where $k$ is the number of possible labels; and the values $0$ and $+1$ represent the irrelevant and relevant label to the instance $x_{i}$ respectively.
\subsection{Label Manifold}
\begin{figure}%[htbp]
\centering
\includegraphics[height=0.35\textheight,width=1\textwidth]{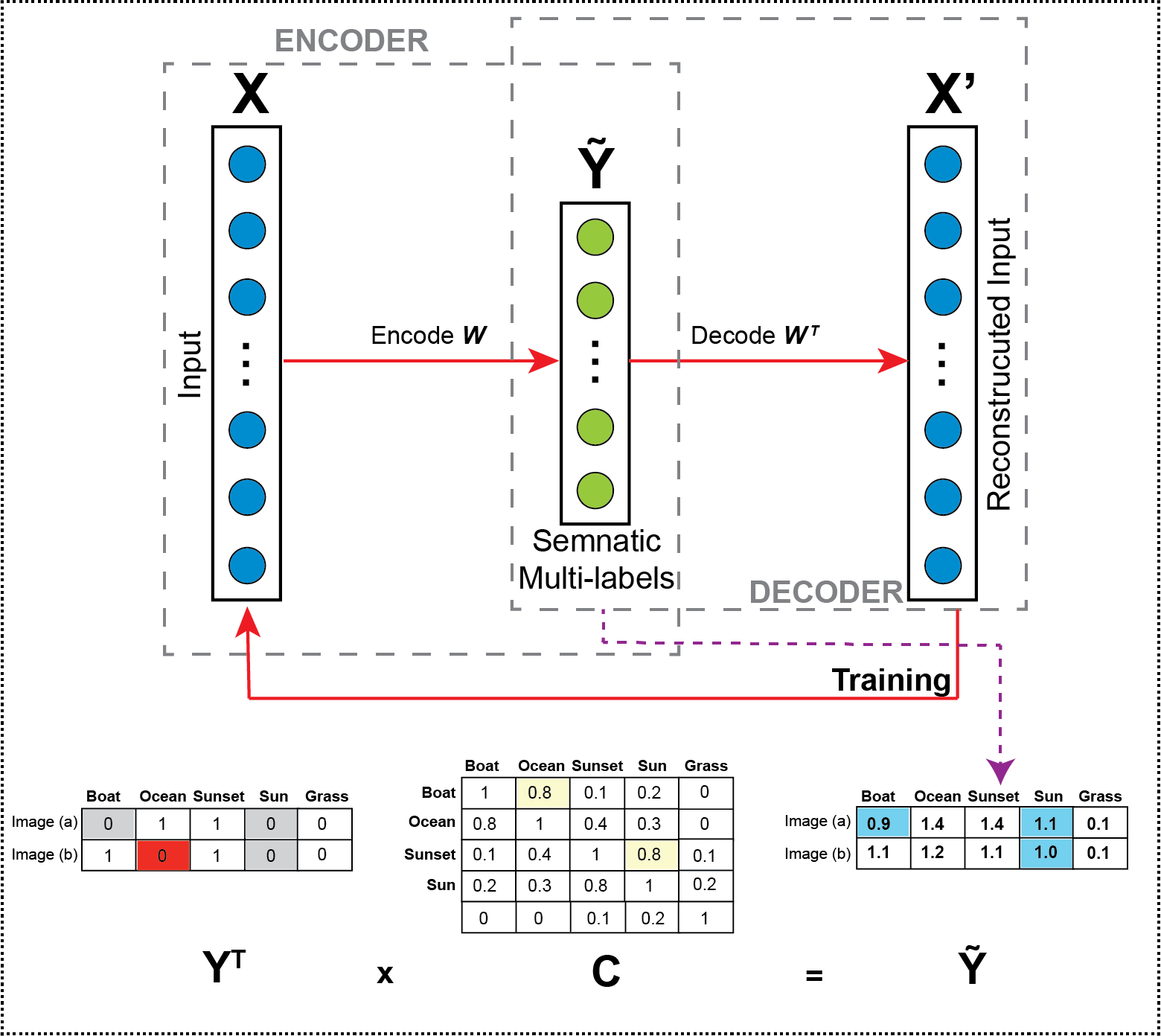}
\caption{Encoder-decoder architecture of our proposed method LDFM. A predicted numerical label matrix $\widetilde{Y}$ is initialized by multiplying the correlation matrix with the original logical label matrix $Y$. The zero values of the images in red, grey and white boxes represent missing, misrepresented, and irrelevant labels, respectively}
\label{fig:correlation}
\end{figure}

To overcome the key challenges in logical label vectors, we first propose to learn a new numerical label matrix $\widetilde{Y} \in \mathbb{R}^{k \times n}$ which contains labels with semantic information. According to the label smoothness assumption~\cite{xu2014learning} which states that if two labels are semantically similar, then their feature vectors should be similar, we initially exploit the dependencies among labels to learn $\widetilde{Y}$ by multiplying the original label matrix with the correlation matrix $C \in \mathbb{R}^{k \times k}$. Due to the existence of logical values in the original label matrix, we use the \textit{Jaccard index} to compute the correlation matrix as follows
%leydesdorff2008normalization
\begin{equation}
\small
\label{eq:newy}
\widetilde{Y}=Y^{T}C 
\end{equation}
where the element $\widetilde{Y}_{i,j}=Y_{i,1}^{T} \times C_{1,j} + Y_{i,2}^{T} \times C_{2,j} + \cdots + Y_{i,n}^{T}\times C_{n,j}$ is initially determined as the predictive numerical value of instance $x_{i}$ is related to the $j$-th label using the prior information of label dependencies. Following is a simple example to investigate the efficiency of using label correlation to learn the semantic numerical labels. The original logical label vectors $Y$ of images (\ref{fig:boat1}) and (\ref{fig:boat2}) are shown in Fig.~\ref{fig:correlation}. The zero values of the original label matrix point to three different types of information. The grey and red colors in Fig.~\ref{fig:correlation} refer to unrepresented and missing labels respectively. However, the white color means that images (\ref{fig:boat1}) and (\ref{fig:boat2}) are not labeled as ``Grass''. We clearly show that the predictive label space $\widetilde{Y}$ can distinguish between three types of zero values and it provides the appropriate numerical values which include semantic information. For example, due to the correlation between ``Boat'' and ``Ocean'' and ``Sunset'' and ``Sun'', the unrepresented label information for ``Boat'' and ``Sun'' in image (\ref{fig:boat1}) and ``Sun'' in image  (\ref{fig:boat2}) is learned. Further, the missing label of the ``Ocean'' image (\ref{fig:boat2}) is predicted. Interestingly, we can further see in the predictive label matrix that the numerical values of the ``Grass'' label in both images are very small because the ``Grass'' label is not correlated with the other labels. Thus, this perfectly matches with the information in the original label matrix that the ``Grass'' object does not exist in images (\ref{fig:boat1}) and (\ref{fig:boat2}) as shown in Fig.~\ref{fig:correlation}. Therefore, based on the above example, we can learn the accurate numerical labels with semantic information. The completed numerical label matrix $\widetilde{Y}$ of the training data is learned in the optimization method of the encoder-decoder framework in the next section.

\subsection{Approach Formulation}
Suppose there is training data $X \in \mathbb{R}^{d \times n}$ with $n$ samples that are associated  with $Y \in \mathbb{R}^{k \times n}$ labels. The predictive numerical matrix $\widetilde{Y}$ is initialised using Eq.~\ref{eq:newy}. The intuition behind our idea is that the proposed method is able to capture the relationship between the feature space and the manifold label space. Inspired by the autoencoder architecture, we develop an effective method that integrates the characteristics of both the low-rank coefficient matrix and semantic numerical label matrix. Specifically, our method is composed of encoder-decoder architecture which tries to learn the projection matrix $W \in \mathbb{R}^{k \times d}$ from the feature space $X$ to the numerical label space $\widetilde{Y}$ in the encoder. At the same time, the decoder can project back to the feature space with $W^{T} \in \mathbb{R}^{d \times k}$ to reconstruct the input training data as shown in Fig.~\ref{fig:correlation}. The objective function is formulated as
%as shown in Fig.~\ref{fig:framework}
\begin{equation}
\small
\label{eq:obj1}
\begin{aligned}
 \min_{W} & \left \| X - W^{T}WX \right \|^{2}_{F}
& \text{s.t. }
  WX=\widetilde{Y}
\end{aligned}
\end{equation}
where $\left \| . \right \|_{F}$ is the Frobenius norm.

%\begin{figure}[ht]
%\centering
%\includegraphics[height=0.15\textheight,width=0.6\textwidth]{Figures/framework.png}
%\caption{Encoder-decoder architecture of our proposed method LDFM. Two-way learning from feature (`F') to semantic label (`S') space (`F' $\rightarrow$ `S' and `S' $\rightarrow$ `F') }
%\label{fig:framework}
%\end{figure}

\subsubsection{Optimization Algorithm}
To optimize the objective function in Eq.~\ref{eq:obj1}, we first substitute $WX$ with $\widetilde{Y}$. Then, due to the existence of constraint $WX=\widetilde{Y}$, it is very difficult to solve Eq.~\ref{eq:obj1}. Therefore, we relax the constraint into a soft one and reformulates the objective function~(\ref{eq:obj1}) as
\begin{equation}
\small
\label{eq:obj2}
\min_{W} \left \| X - W^{T}\widetilde{Y} \right \|^{2}_{F}  + \lambda \left \| WX - \widetilde{Y}\right \|^{2}_{F}
\end{equation}
where $\lambda$ is a parameter to control the importance of the second term. Now, the objective function in Eq.~\ref{eq:obj2} is non-convex and it contains two unknown variables $W$ and $\widetilde{Y}$. It is difficult to directly solve the equation. We propose a solution to iteratively update one variable while fixing the other. Since the objective function is convex by updating one variable, we compute the partial derivative of Eq.~\ref{eq:obj2} with respect to $W$ and $\widetilde{Y}$ and set both to zero.
\begin{algorithm}%[htbp]
{\small
    \KwIn{Training data $X \in \mathbb{R}^{d \times n}$}
    \myinput{Logical label matrix $Y \in \mathbb{R}^{k \times n}$}
     \myinput{Parameters: $\lambda$ and MaxIteration}
     
     \textbf{Initialization:}\\
      $C_{jk}=\frac{\sum_{i=1}^{n}x_{ij}x_{jk}}{\sum_{i=1}^{n}x_{ij} + \sum_{i=1}^{n}x_{ik} - \sum_{i=1}^{n}x_{ij}x_{ik}}; j,k$ are the label vectors\\
      $\widetilde{Y}=YC$\\
      $t=0$
   
    % STEP 1
    \While{MaxIteration$>t$}{
        % STEP 2
          Update $W$ by the solving the Eq.~\ref{eq:sylv_W}\;
          Update $\widetilde{Y}$ by the solving the Eq.~\ref{eq:sylv_Y}\;

           $t= t + 1$\;
    }
    
    \KwOut{Projection matrix $W \in \mathbb{R}^{k \times d}$}
    Rank features by $\left \| W_{m,:} \right \|_{2}$ in descending order and return the top ranked features
    \caption{Learning Discriminative Features using Multi-label Dual Space (LDFM)}
    %\caption{LDFM method}
 
  \label{mainalgo} 
  }
\end{algorithm}  

\begin{itemize}
	\item \textbf{Update $W$:}
	\begin{equation}
	\small
\label{eq:W}
\begin{aligned}
 & -\widetilde{Y}\left ( X^{T} - \widetilde{Y}^{T}W \right ) + \lambda\left ( WX-\widetilde{Y} \right )X^{T}=0 \\
 & \Rightarrow \widetilde{Y} \widetilde{Y}^{T}W + \lambda WXX^{T}=\widetilde{Y}X^{T} + \lambda \widetilde{Y} X^{T} \\
 & \Rightarrow PW + WQ=R
\end{aligned}
\end{equation}
where $P=\widetilde{Y}\widetilde{Y}^{T}$, $Q=\lambda XX^{T}$, and $R=\left ( \lambda + 1 \right )\widetilde{Y}X^{T}$ 
	
\item \textbf{Update $\widetilde{Y}$:}
\begin{equation}
\small
\label{eq:Y}
\begin{aligned}
 & -WX+WW^{T}\widetilde{Y} + \lambda\left ( -WX + \widetilde{Y} \right )=0 \\
 & \Rightarrow WW^{T}\widetilde{Y} + \lambda \widetilde{Y} = WX + \lambda WX \\
 & \Rightarrow A\widetilde{Y}+\widetilde{Y}B=D
\end{aligned}
\end{equation}
where $A=WW^{T}$, $B=\lambda I$, $D=\left ( \lambda + 1 \right )WX$ and $I \in \mathbb{R}^{k \times k}$ is an identity matrix.
\end{itemize}
Eqs.~\ref{eq:W} and \ref{eq:Y} are formulated as the well-known Sylvester equation of the form $MX+XN=O$. The sylvester equation is a matrix equation with given matrices $M, N, $ and $O$ and it aims to find the possible unknown matrix $X$. The solution of the Sylvester equation can be solved efficiently and lead to a unique solution. For more explanations and proofs, the reader can refer to~\cite{bartels1972solution}. Using the Kronecker products notation and the vectorization operator $vec$, Eqs.~\ref{eq:W} and \ref{eq:Y} can be written as a linear equation respectively

\begin{equation}
\small
\label{eq:sylv_W}
\left (I_{d} \otimes P + Q^{T} \otimes I_{k} \right )vec(W)=vec(R)
\end{equation}
where $I_{d} \in \mathbb{R}^{d \times d}$ and $I_{k} \in \mathbb{R}^{k \times k}$ are identity matrices and $\otimes$ is the Kronecker product.

\begin{equation}
\small
\label{eq:sylv_Y}
\left (I_{n} \otimes A + B^{T} \otimes I_{k} \right )vec(\widetilde{Y})=vec(D)
\end{equation}
where $I_{n} \in \mathbb{R}^{n \times n}$ is an identity matrix. Fortunately, in MATLAB, this equation can be solved with a single line of code \textit{sylvester}\footnote{https://uk.mathworks.com/help/matlab/ref/sylvester.html}.
Now, the two unknown matrices $W$ and $\widetilde{Y}$ can be iteratively updated using the proposed optimization rules above until convergence. The procedure is described in Algorithm~\ref{mainalgo}.

Our proposed method learns the encoder projection matrix $W$. Thus, we can embed a new test sample $x^{s}_{i}$ to the semantic label space by $\widetilde{y}_{i}=Wx^{s}_{i}$. Similarly, we can reconstruct the original features using the decoder projection matrix $W^{T}$ by $x^{s}_{i}=W^{T}\widetilde{y}_{i}$. Therefore, $W$ contains the discriminative features to predict the real semantic labels. To identify these features, we rank each feature according to the value of $\left \| W_{m,:} \right \|_{2} \left ( m=1,\cdots,d \right )$ in descending order and return the top ranked features.

\begin{table}%[t]
\centering
{\scriptsize
\begin{tabular}{llccccl}
\hline
Dataset & Domain & \#Instance & \#Training & \#Test & \#Features & \#Labels  \\
\hline
Scene &Image & $2407$&$1211$&$1196$&$294$&$6$ scenes\\
Emotions &Audio &  $593$&$391$&$202$&$72$&$6$ emotions\\
Reference  &Text (Yahoo) & $5000$&$2000$&$3000$&$793$&$33$ topics\\
Computers & Text (Yahoo)&  $5000$&$2000$&$3000$&$681$&$33$ topics\\
\hline
\end{tabular}
}
\caption{Characteristics of the evaluated datasets}
\label{tbl:ds}
\end{table}

\section{Experiments} 
\subsection{Experimental datasets}
We open source for our LDFM code for reproducibility of our experiments\footnote{https://github.com/alibraytee/LDFM}. Experiments are conducted on four public multi-label datasets which can be downloaded from the Mulan repository\footnote{http://mulan.sourceforge.net/datasets-mlc.html}. The details of these datasets are summarized in Table~\ref{tbl:ds}. Due to space limitations, we use three evaluation metrics namely \textit{Hamming loss}, \textit{Average precision}, and \textit{Micro-F1} which define in~\cite{braytee2019correlated}. 
%Among these datasets, Scene contains 2407 images where each image is associated with six scenes. Emotions contains 593 songs related to six emotions, and on text, Reference and Computers are used from Yahoo datasets
\subsection{Comparing methods and experiment settings}
%As recommended by~\cite{he2005face}, 
Multi-label feature selection methods attracted the interest of the researchers in the last decade. In this study, we compare our proposed method LDFM against the recent state-of-the-art multi-label feature selection methods, including GLOCAL~\cite{zhu2017multi}, MCLS~\cite{huang2018manifold}, and MSSL~\cite{cai2018multi}. MCLS and MSSL methods consider feature manifold learning in their studies. ML-KNN ($K$=$10$)~\cite{zhang2007ml} is used as the multi-label classifier to evaluate the performance of the identified features. The results based on a different number of features, vary from $1$ to $100$ features. PCA is applied as a prepossessing step with and retain $95\%$ of the data. To ensure a fair comparison, the parameters of the compared methods are tuned to find the optimum values. For GLOCAL, the regularization parameters $\lambda_{3}$ and $\lambda_{4}$ are tuned in $\left \{ 0.0001,0.001,\dots, 1 \right \}$, the number of clusters is searched from $\left \{ 4,8,16,32, 64 \right \}$, and the latent dimensionality rank is tuned in $\left \{ 5,10,\dots, 30 \right \}$. MCLS uses the defaults settings for its parameters. For MSSL, the parameters $\alpha$ and $\beta$ are tuned by searching the grid $\left \{ 0.001,0.01,\dots,1000 \right \}$. The parameter settings for LDFM are described in the following section.
\begin{figure}[ht]
%\centering
\includegraphics[height=0.15\textheight,width=1\textwidth]{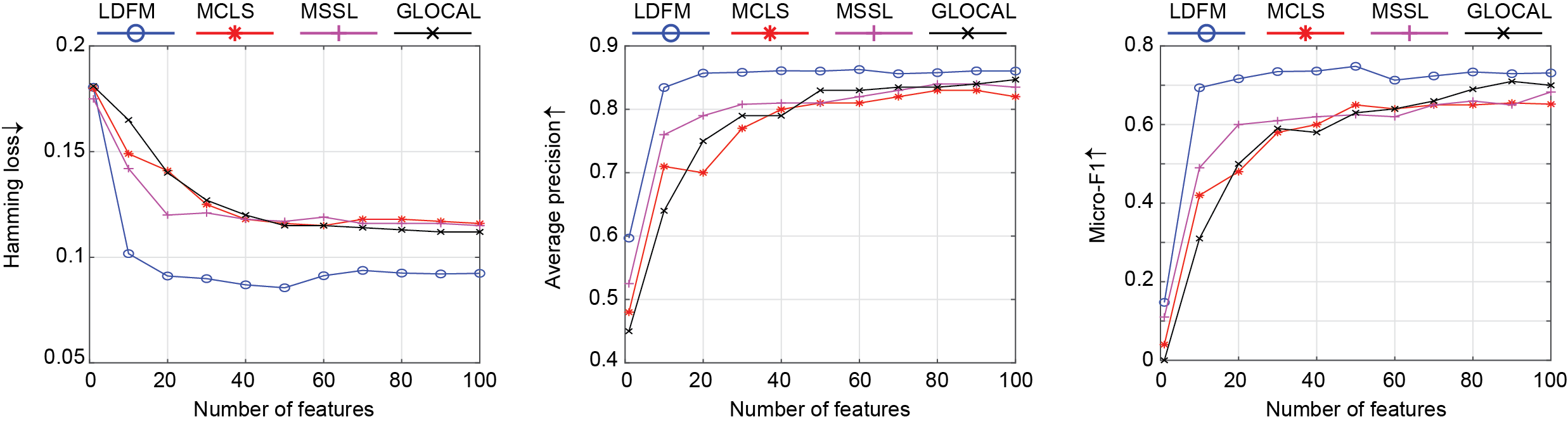}
\caption{Comparison of multi-label feature selection algorithms on Scene}
\label{fig:scene}
\end{figure}

\begin{figure}[ht]
%\centering
\includegraphics[height=0.15\textheight,width=1\textwidth]{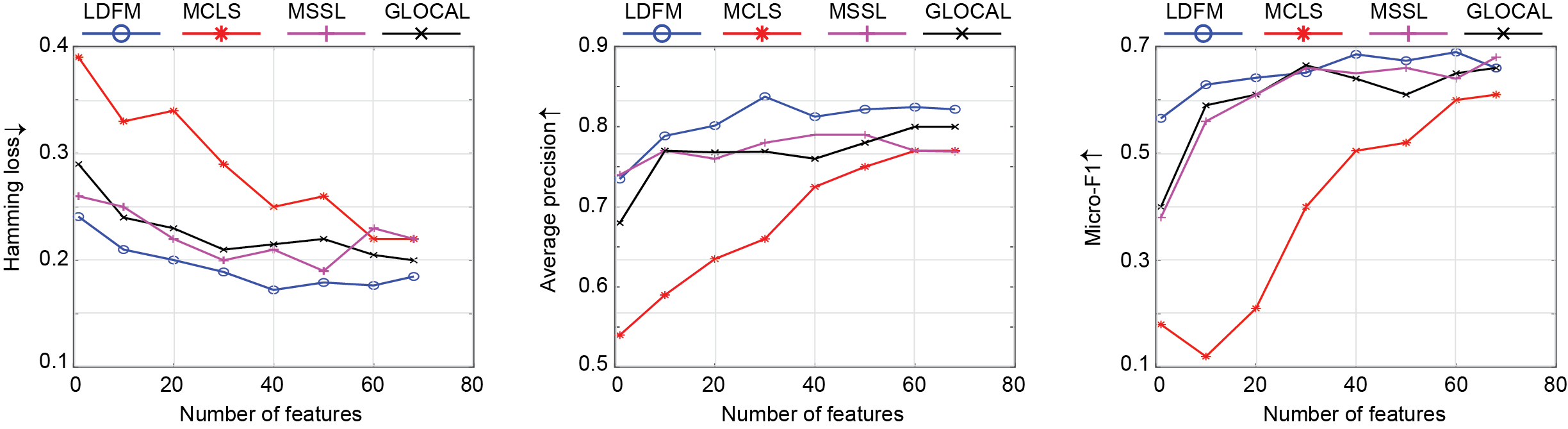}
\caption{Comparison of multi-label feature selection algorithms on Emotions}
\label{fig:emotions}
\end{figure}
\subsection{Results}
\subsubsection{Classification results}
\begin{figure}[ht]
%\centering
\includegraphics[height=0.15\textheight,width=1\textwidth]{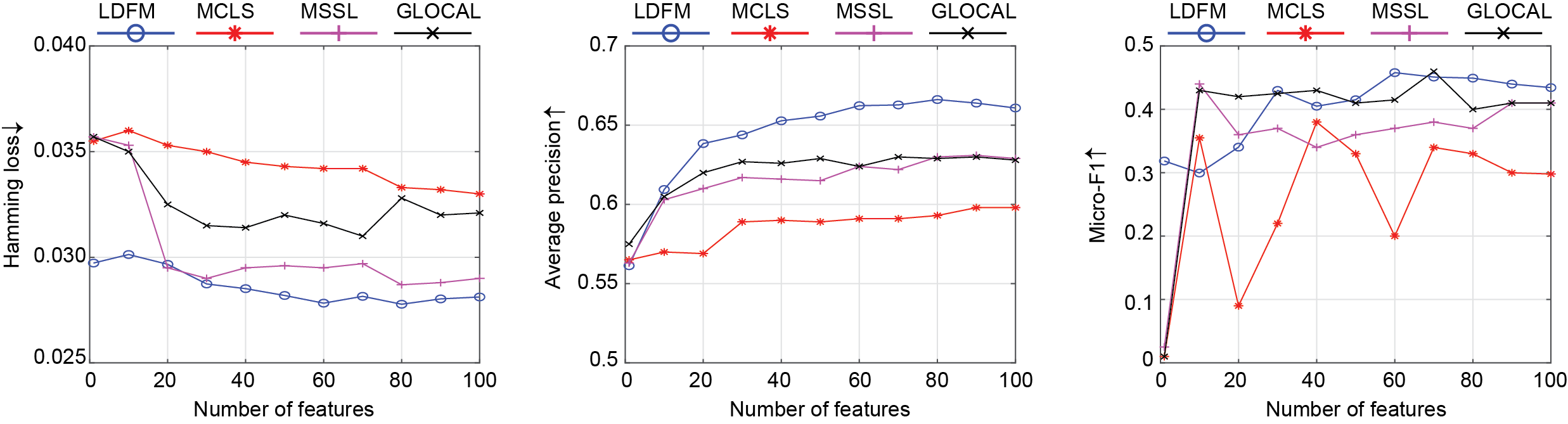}
\caption{Comparison of multi-label feature selection algorithms on Reference}
\label{fig:reference}
\end{figure}

\begin{figure}[ht]
%\centering
\includegraphics[height=0.15\textheight,width=1\textwidth]{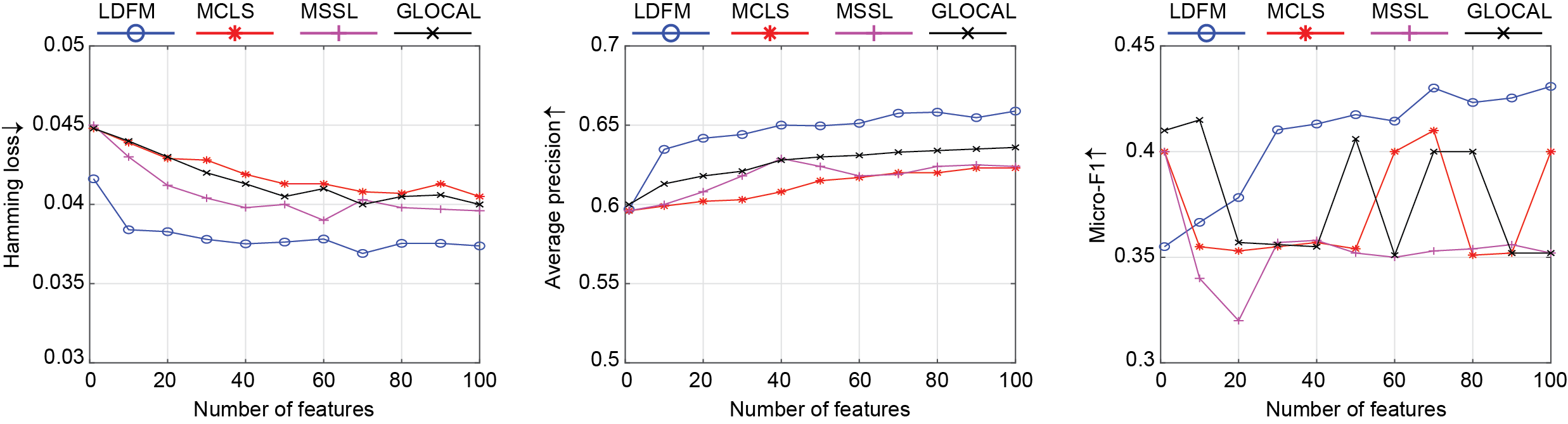}
\caption{Comparison of multi-label feature selection algorithms on Computers}
\label{fig:computers}
\end{figure}

Several experiments have been conducted to demonstrate the classification performance of LDFM compared to the state-of-the-art multi-label feature selection methods. Figs.~\ref{fig:scene}-\ref{fig:computers} show the results in terms of \textit{Hamming loss}, \textit{Average precision}, and \textit{Micro-F1} evaluation metrics on the four datasets. In these figures, the classification results are generated based on top-ranked $100$ features (except Emotions which only has $72$ features). Based on the experiment results shown in Figs.~\ref{fig:scene}-\ref{fig:computers}, interestingly, it is clear that our proposed method has a significant classification improvement with an increasing number of selected features, and then remains stable. Thus, this observation indicates that it is meaningful to study dimensionality reduction in multi-label learning. Further, it highlights the stability and capability of LDFM to achieve good performance on all the datasets with fewer selected features.

The proposed method is compared with the state-of-the-art on each dataset. As shown in Figs.~\ref{fig:scene}-\ref{fig:computers}, LDFM achieves better results compared to MCLS, MSSL, and GLOCAL on almost all the evaluated datasets. Specifically, in terms of the \textit{Hamming loss} evaluation metric, where the smaller the values, the better the performance, LDFM's features substantially improve the classification results compared to the state-of-the-art. It can be observed that MCLS has the worst results and MSSL and GLOCAL achieve comparable results as shown in Figs.~\ref{fig:scene}, \ref{fig:emotions}, and \ref{fig:computers}. In terms of \textit{Average precision}, and \textit{Micro-F1} evaluation metrics, where the larger the values, the better the performance, LDFM generally achieves better results against the compared methods in all datasets. We note that LDFM performs slightly better than MSSL and GLOCAL on the Emotions dataset using \textit{Micro-F1} metric. We also note that the compared methods produce unstable results on the Reference and Computers datasets using the \textit{Micro-F1} metric. In general, our proposed method demonstrates the great benefits of using label manifolds in encoder-decoder architecture to identify the discriminative features. Furthermore, using the Friedman test, we investigate whether the results that are produced by LDFM are significantly different to the state-of-the-art. In particular, we examine the Friedman test between LDFM against each compared method for each evaluation metric in the four datasets. The statistical results show that the p-value in all tests is less than $0.05$ which rejects the null hypothesis that the proposed method and the compared methods have an equal performance. Finally, we explored the reconstruction capability of the decoder in LDFM using the projection matrix $W$ to reconstruct the original data. Table~\ref{tbl:reconst} reports the reconstruction errors using the training logical labels ($Y$), training predicted labels ($\widetilde{Y}$), and the logical testing labels. It is observed that the percentage reconstruction error of the original training matrix using the logical training labels only ranges between $4\%$ to $8\%$ on the four dataset, and this error is dramatically decreased to $0.1\%$ to $3\%$ by using the predicted numerical label matrix. This observation reveals that the decoder plays an important role in selecting the important features which can be used to reconstruct the original matrix. Further, it supports our argument to reconstruct the visual images using the semantic labels and the coefficient matrix. In addition, we report the capability of reconstructing the testing data matrix using the projection matrix and the testing labels with a small error that ranges between $4\%$ to $8\%$.

\begin{table}%[t]
\centering
\scriptsize
\begin{tabular}{lccc}
\hline
Dataset & Logical error & Predicted error & Testing error \\
\hline
Scene & $0.042$&$0.032$ & $0.043$  \\
Emotions & $0.087$&$0.022$ & $0.086$ \\
Reference & $0.044$&$0.001$&$0.044$   \\
Computers & $0.051$&$0.001$&$0.051$   \\
\hline
\end{tabular}
\caption{Reconstruction different error values using the decoder}
\label{tbl:reconst}
\end{table}

\subsubsection{LDFM results for handling missing labels}
In this experiment, to investigate the ability of the proposed method to handling missing labels, we randomly removed different proportions of labels from the samples from moderate to extreme levels: 20\%, 40\%, 60\%, and 80\%. Table~\ref{tblmissing} shows that LDFM achieved consistent improvement over the base especially on 20\%, 40\%, and 60\% missing label levels. Further, it is superior across four datasets using four different missing label proportions. %The success of LDFM due to learning the semantic label matrix using the original label matrix as shown in Fig.~\ref{fig:correlation}.
\begin{table}[t]
\centering
\scriptsize
\begin{tabular}{l|l|llll|llll}
\hline
 Dataset$\downarrow$  & Evaluation criteria & \multicolumn{4}{c|}{Base } & \multicolumn{4}{c}{\textbf{LDFM} } \\ 
%\hline
 & Missing label proportion $\rightarrow$ & 20\% & 40\% & 60\% & 80\% & 20\% & 40\% & 60\% & 80\% \\ 
\hline
\multirow{3}{*}{Scene} & Hamming Loss & 0.18 & 0.24 & 0.69 & 0.81 & 0.10 & 0.14 & 0.39 & 0.78 \\
 & Average Precision & 0.57 & 0.54 & 0.51 & 0.49 & 0.81 & 0.79 & 0.76 & 0.73 \\
 & Micro-F1 & 0.25 & 0.32 & 0.32 & 0.30 & 0.67 & 0.63 & 0.44 & 0.32 \\ 
\hline
\multirow{3}{*}{Computers} & Hamming Loss & 0.04 & 0.05 & 0.18 & 0.91 & 0.03 & 0.04 & 0.17 & 0.91 \\
 & Average Precision & 0.58 & 0.56 & 0.55 & 0.55 & 0.61 & 0.59 & 0.57 & 0.56 \\
 & Micro-F1 & 0.39 & 0.40 & 0.24 & 0.09 & 0.42 & 0.42 & 0.26 & 0.10 \\ 
\hline
\multirow{3}{*}{Reference} & Hamming Loss & 0.03 & 0.04 & 0.15 & 0.92 & 0.02 & 0.03 & 0.14 & 0.91 \\
 & Average Precision & 0.54 & 0.52 & 0.50 & 0.50 & 0.60 & 0.57 & 0.55 & 0.54 \\
 & Micro-F1 & 0.40 & 0.41 & 0.20 & 0.07 & 0.42 & 0.45 & 0.23 & 0.07 \\ 
\hline
\multirow{3}{*}{Emotions} & Hamming Loss & 0.23 & 0.33 & 0.57 & 0.65 & 0.21 & 0.28 & 0.51 & 0.64 \\
 & Average Precision & 0.76 & 0.74 & 0.70 & 0.68 & 0.79 & 0.76 & 0.75 & 0.72 \\
 & Micro-F1 & 0.62 & 0.59 & 0.50 & 0.48 & 0.65 & 0.63 & 0.54 & 0.49 \\
\hline
\end{tabular}
\caption{Results on four datasets with different missing label proportions}
\label{tblmissing}
\end{table}

\subsection{Parameter sensitivity and convergence analysis}
\label{sec:param}
\begin{figure}[htbp]
%\centering
\subfloat[Emotions]{% 
\includegraphics[height=0.1\textheight,width=0.25\textwidth]{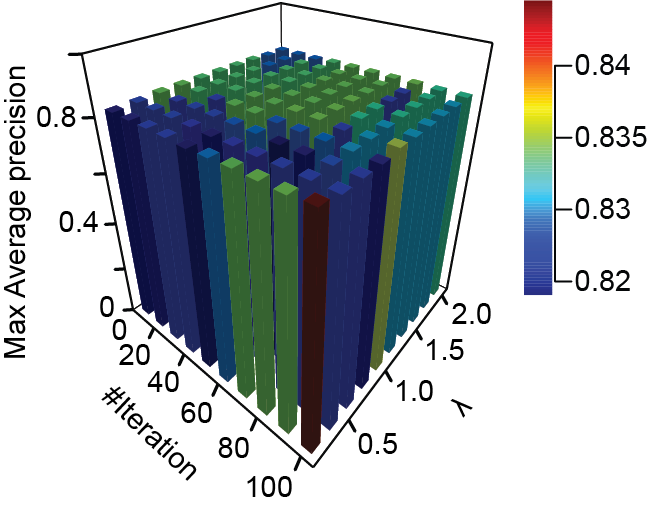}
} 
%\hfill
\subfloat[Reference]{% 
\includegraphics[height=0.1\textheight,width=0.25\textwidth]{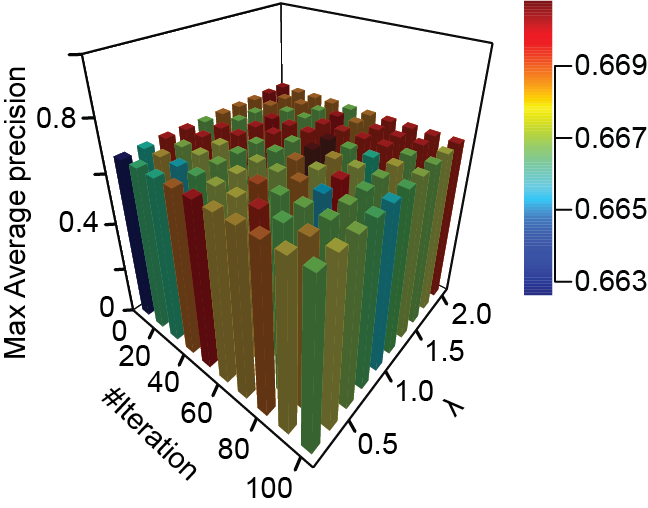} %label{fig:2} 
}
\hfill
\subfloat[Emotions]{% 
\includegraphics[height=0.1\textheight,width=0.2\textwidth]{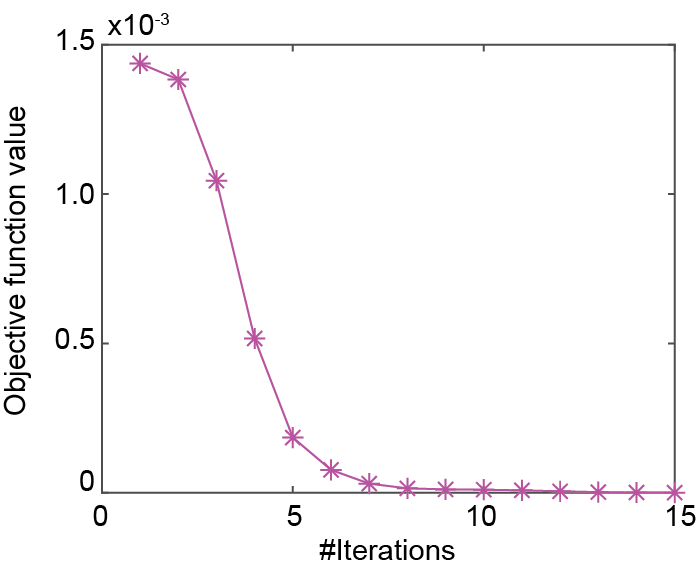}
}
\subfloat[Reference]{% 
\includegraphics[height=0.1\textheight,width=0.2\textwidth]{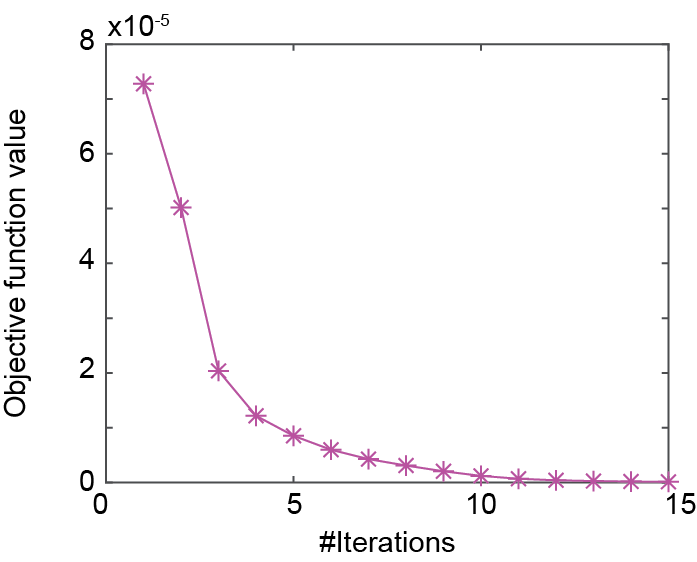}
}

\caption{LDFM Results on Emotions and Reference datasets. (a) and (b) the average precision results w.r.t different parameters. (c) and (d) convergence curves} 
\label{fig:param}
\end{figure}

In this section, we study the influence of the proposed method's parameters $\lambda$ and $MaxIteration$ on the classification results. First, $\lambda$ controls the contribution of the decoder and encoder in the method, however the second parameter defines the number of iterations required to convergence. The parameter $\lambda$ and $MaxIteration$ are tuned using a grid search from $\left \{ 0.2,0.4,\dots,2 \right \}$ and $\left \{ 1,20,40,\dots,100 \right \}$ respectively. As shown in Fig.~\ref{fig:param}a and \ref{fig:param}b, using the average precision metric on the two datasets, we can observe that with an increasing $\lambda$, the learning performance is improved. Further, we investigate the convergence of the LDFM optimization method. As shown in Fig.\ref{fig:param}c and \ref{fig:param}d using two datasets, it is clearly seen that our method converges rapidly and has around $10$ iterations which demonstrates the efficacy and speed of our algorithm.

%\section{Conclusion}
%This paper proposes a novel semantic multi-label learning model based on an autoencoder. Most existing multi-label learning methods learn feature encoding to map the original training examples to the logical labels. But, in addition to feature encoding our proposed method explores the reconstruction of the original data using numerical semantic labels. The semantic labels are predicted in the optimization method because they are not explicitly available from the training samples. We further rank the feature weights in the learned project matrix for feature selection. The proposed method is simple and computationally fast. We demonstrate through extensive experiments that our method outperforms the state-of-the-art. Furthermore, we demonstrate the efficiency of the proposed method to reconstruct the original data using the predicted labels.
\section{Conclusion}
This paper proposes a novel semantic multi-label learning model based on an autoencoder. Our proposed method learns the projection matrix to map from the feature space to semantic space back and forth. The semantic labels are predicted in the optimization method because they are not explicitly available from the training samples. We further rank the feature weights in the learned project matrix for feature selection. The proposed method is simple and computationally fast. We demonstrate through extensive experiments that our method outperforms the state-of-the-art. Furthermore, we demonstrate the efficiency of the proposed method to reconstruct the original data using the predicted labels.

\bibliographystyle{splncs04}
\bibliography{LDFM}
\end{document}